\documentclass[conference]{IEEEtran}
\IEEEoverridecommandlockouts
\usepackage{cite}
\usepackage{amsmath,amssymb,amsfonts}
\usepackage{algorithmic}
\usepackage{algorithm}
\usepackage{graphicx}
\usepackage{textcomp}
\usepackage{xcolor}
\usepackage{bm}
\renewcommand{\baselinestretch}{0.925}
\def\BibTeX{{\rm B\kern-.05em{\sc i\kern-.025em b}\kern-.08em
    T\kern-.1667em\lower.7ex\hbox{E}\kern-.125emX}}
\begin{document}

\title{Bayesian Optimization Approach for Analog Circuit Synthesis Using Neural Network \\
%{\footnotesize \textsuperscript{*}Note: Sub-titles are not captured in Xplore and should not be used}
%\thanks{Shuhan Zhang, Wenlong Lyu, Fan Yang, Changhao Yan and Xuan Zeng are with the State Key Laboratory of ASIC and System, School of Microelectronics, Fudan University, Shanghai 201203, China. Dian Zhou is with the Department of Electrical Engineer, University of Texas at Dallas, USA. (email: yangfan@fudan.edu.cn, yanch@fudan.edu.cn)}
\thanks{* Corresponding authors: \{yangfan, yanch\}@fudan.edu.cn}
}

\author{Shuhan Zhang$^1$, Wenlong Lyu$^1$, Fan Yang$^1$, Changhao Yan$^1$, Dian Zhou$^{1,2}$, Xuan Zeng$^1$   \\
   \normalsize{ $^1$State Key Lab of ASIC \& System, Microelectronics Department, Fudan University, China}  \\
   \normalsize{ $^2$Department of Electrical Engineering, University of Texas at Dallas, USA}
}

\maketitle

\begin{abstract}
Bayesian optimization with Gaussian process as surrogate model has been successfully applied to analog circuit synthesis. In the traditional Gaussian process regression model, the kernel functions are defined explicitly. The computational complexity of training is $O(N^3)$, and the computation complexity of prediction is $O(N^2)$, where $N$ is the number of training data. Gaussian process model can also be derived from a weight space view, where the original data are mapped to feature space, and the kernel function is defined as the inner product of nonlinear features. In this paper, we propose a Bayesian optimization approach for analog circuit synthesis using neural network. We use deep neural network to extract good feature representations, and then define Gaussian process using the extracted features. Model averaging method is applied to improve the quality of uncertainty prediction. Compared to Gaussian process model with explicitly defined kernel functions, the neural-network-based Gaussian process model can automatically learn a kernel function from data, which makes it possible to provide more accurate predictions and thus accelerate the follow-up optimization procedure. Also, the neural-network-based model has $O(N)$ training time and constant prediction time. The efficiency of the proposed method has been verified by two real-world analog circuits.
\end{abstract}


\section{Introduction}
Although the digital circuit designs have been automated for a long time, most of the analog circuits are still manually designed. Due to the more and more complicated device models and the serve process variations, manual analog circuit designs become more and more challenging. The growing demand for the high performance, low power and time to market make automated analog circuit design a necessity~\cite{Rutenbar2007Hierarchical}. Automated analog circuit design has attracted many interests in both industry and academic community.

The automated circuit design includes two ingredients: the topology selection and the device sizing. We concentrate on the device sizing in this paper. Generally, the traditional sizing algorithms can be divided into two categories: the model-based and simulation-based approaches\cite{Lyu2017An}. The model-based approaches try to generate simplified or regression models which represent the performances of the circuits. Geometric programming\cite{Boyd2007A} based method uses posynomials to model the performances. The sizing problem can be formulated as a convex programming problem, and thus the global optimum can be found efficiently. General polynomials\cite{wang2014enabling} can also be used to model the performances, and the general polynomial programming\cite{wang2014enabling} problems can be transformed to convex programming problems through Semidefinite-Programming (SDP) relaxations. One of the main difficulties of the model-based approaches is that the models would deviate from the real circuit performances. The obtained global optimums would also deviate from the real optimums.

Simulation-based algorithms take the objective functions and constraints as the black-box functions. They employ optimization strategies to propose new candidates, and aim to better explore the design space to search for the global optimum. The corresponding optimization approaches include simulated annealing~\cite{Phelps2006Anaconda}, particle swarm intelligence approach~\cite{Vural2012Analog}, evolutionary algorithm~\cite{liu2009analog}, and multi-start optimization algorithm~\cite{Peng2016Efficient,lv2017subgradient,Yang2018Smart}. Since the simulations are performed on-the-fly, the number of simulations would even be lower than the model-based approaches. However, the convergence rates of these simulation-based methods are relatively low due to corresponding optimization strategies.

Recently, surrogate-model based optimization methods have been proposed for device sizing. The optimizations depend on surrogate models, which are evolved gradually during the optimization process. Gaussian Process Regression (GPR) model can provide the predictions as well as the uncertainties of the predictions, and has been widely used as surrogate model in these methods. In GASPAD method~\cite{Liu2014GASPAD}, the GPR model combined with evolutionary algorithm is proposed for the optimization of microwave circuits. In WEIBO method~\cite{Lyu2017An}, GPR model based Bayesian optimization algorithm was proposed. Acquisition functions are introduced in Bayesian optimization to balance the exploration and exploitation. The Expected Improvement (EI) weighted by the probability of satisfying the constraints is selected as the acquisition functions in~\cite{Lyu2017An} to handle the constrained optimization problems in device sizing. It was reported that the WEIBO method can achieve better optimization results with significantly less number of simulations than GASPAD method~\cite{Lyu2017An}.

In this paper, we propose a Bayesian optimization approach for analog circuit synthesis using neural network. In the traditional Gaussian process regression model, the kernel functions are defined explicitly. The computational complexity of training is $O(N^3)$, and the computational complexity of prediction for the traditional model is $O(N^2)$, where $N$ is the number of training data. Gaussian process model can also be derived from a weight space view, where the original data are mapped to feature space, and the kernel functions are implicitly defined in the feature space. We propose a well-designed neural network to provide efficient feature representations. Model averaging is proposed to improve the quality of uncertainty prediction. Compared to Gaussian process regression model with explicitly defined kernel functions, neural-network-based models are possible to provide more accurate predictions and thus accelerate the follow-up optimization procedure. Moreover, the neural-network-based model has $O(N)$ training time and constant prediction time. The effectiveness of our approach is demonstrated by real-world analog circuits.

The rest of the paper is organized as follows. In section 2, we review the background of device sizing, the GPR model and Bayesian optimization algorithm. In section 3, we present our proposed Bayesian optimization approach using neural network. In section 4, we demonstrate the efficiency of the proposed approach using real-world analog circuits. We conclude the paper in section 5.

\section{Background}
In this section, we will review the device sizing problem, Bayesian optimization framework and the Gaussian process regression model.

\subsection{Problem Formulation of Device Sizing}
The device sizing problem can be formulated as a constrained optimization problem.
\begin{equation}
    \label{eq:Formulation}
    \begin{aligned}
        & \text{minimize} & & f(\bm{x}) \\
        & \text{s.t.}     & & g_i(\bm{x}) < 0 \\
        &                 & & \forall i \in {1 \dots N_c},
    \end{aligned}
\end{equation}
where $\bm{x} \in R^d$ represents $d$ design variables,  $f(\bm{x})$ denotes
the Figure of Merit (FOM) of the analog circuit. Without loss of generality, we only consider the minimization problem and the constraints are formulated as
$g_i(\bm{x}) < 0, \forall i \in {1 \dots N_c}$, where a $g_i(\bm{x}) < 0$
corresponds to the $i$-th constraints.

\subsection{Overview of Bayesian Optimization}
Bayesian optimization is a framework for model-based global optimization of noisy and expensive black-box functions~\cite{Shahriari2015Taking}. Generally, it consists of a probabilistic surrogate model and an acquisition function which helps to trade off exploration and exploitation. The probabilistic surrogate model provides the prediction and well-calibrated uncertainty. The probabilistic surrogate model, which captures our beliefs about the behavior of the unknown objective function. When a new data point is observed, the surrogate model is updated to generate a more informative posterior distribution. The balance of exploration and exploitation is very important for an optimization algorithm. The exploration means the algorithm try to explore the state space where the uncertainty predicted by surrogate model is high. The exploitation means the algorithm tends to find the optimum predicted by surrogate model with high probability. There are several typical acquisition functions, e.g., expected improvement (EI)~\cite{Mockus1978The}, upper confidence bounds (UCB)~\cite{auer2002using}.

In Bayesian optimization, a surrogate model is built firstly based on a initial training set which is randomly sampled. In each iteration, a new data point is obtained by maximizing the acquisition function. The surrogate models are then updated using the new data point for the next iteration.

\subsection{Gaussian Process Regression}
The surrogate model widely used in Bayesian optimization is the Gaussian process (GP) regression model. Given a training set $D_{T}=\{X,\bm{y}\}$ where $X = \{\bm{x}_{1},\bm{x}_{2},~\dots~,\bm{x}_{N}\}$, and $\bm{y} = \{y_{1},y_{2},~\dots~,y_{N}\}$, we assume that $\bm{y}$ is generated by a black-box function $f(\bm{x})$ with additive noise $\epsilon \sim N(0,\sigma_n^{2})$, 
\begin{equation}
    y_{i} \sim N(f(\bm{x}_{i}),\sigma_n^{2}).
\end{equation}
Here, $N(.,.)$ denotes a Gaussian distribution. For a Gaussian process, it assumes that any finite samples follows a joint Gaussian distribution $\bm{f} \sim N(\bm{m},K)$, where $\bm{m}$ is the mean vector $\bm{m} = (m(\bm{x}_{1}),m(\bm{x}_{2}),~\dots~,m(\bm{x}_{N}))^{T}$, and $K_{i,j} = k(\bm{x}_{i},\bm{x}_{j})$ is an element of the covariance matrix. The mean function $m(\bm{x})$ can be any function, while the kernel function $k(\bm{x}_i,\bm{x}_j)$ has to make sure that the covariance matrix is a symmetric positive definite (SPD) matrix. In \cite{Lyu2017An}, constant mean function $m(\bm{x})=\mu_0$ and the Gaussian kernel function 
$$
k(\bm{x}_i, \bm{x}_j) = \sigma_n^2 \exp(-\frac{1}{2} (\bm{x}_i - \bm{x}_j)^T \Lambda^{-1} (\bm{x}_i - \bm{x}_j))
$$
are used, where $\mu_0$, $\sigma_n$ and $\Lambda = \text{diag}(l_1, l_2,~\dots,~l_d)$ are hyper parameters.

Given a new input vector $\bm{x}$, GPR model predicts the distribution of $y \sim N(\mu(\bm{x}),\sigma^2(\bm{x}))$. The $\mu(\bm{x})$ and $\sigma^2(\bm{x})$ can be expressed as~\cite{Lyu2017An}
\begin{equation}
    \label{eq:mu_and_sigma}
    \begin{cases}
	\mu(\bm{x})= \mu_0 + k(\bm{x},X)(K+\sigma^{2}_{n}I)^{-1}(\bm{y}-\mu_0) \\
        \sigma^{2}(\bm{x})=\sigma^{2}_{n}+k(\bm{x},\bm{x})-k(\bm{x},X)(K+\sigma^{2}_{n}I)^{-1}k(X,\bm{x}),
    \end{cases}
\end{equation}
where $k(\bm{x},X) = (k(\bm{x},\bm{x}_{1}),k(\bm{x},\bm{x}_{2}),~\dots~,k(\bm{x},\bm{x}_{N}))$ and $k(X,\bm{x})=k(\bm{x},X)^{T}$. In (\ref{eq:mu_and_sigma}), $\mu(\bm{x})$ can be viewed as the prediction, and $\sigma^{2}(\bm{x})$ can be viewed the confidence of the prediction. Denote $\bm{\theta}$ as the vector of hyper parameters including the hyper parameters for the kernel function, and the additive noise. The hyper parameters can be learned by maximum likelihood estimation (MLE) based on the training data~\cite{Rasmussen2005Gaussian}
\begin{equation}
    \label{eq:MLE}
    \log p(\bm{y}|X,\bm{\theta}) = -\frac{1}{2}(\bm{y}^{T} K_{\bm{\theta}}^{-1} \bm{y}+\log \lvert K_{\bm{\theta}} \rvert + N \log(2\pi)),
\end{equation}
where $K_{\bm{\theta}}$ is the covariance matrix of the training input calculated by the kernel function, and $K_{\bm{\theta}}=K+\sigma^{2}_{n}I$.

\subsection{Acquisition Function}
The acquisition function of Bayesian optimization helps to balance between the exploration and exploitation. One widely used acquisition function is the expected improvement (EI) criterion~\cite{Mockus1978The}. Let $\tau$ denotes the current minimum objective value. The improvement is calculated as follows:
\begin{equation}
    I(y) = \\
    \begin{cases}
        0 & y >= \tau \\
        \tau - y & y < \tau .
    \end{cases}
\end{equation}
Here, $I(y)$ denotes the improvement over the target value $\tau$. Considering the distribution provided by the probabilistic surrogate model $y \sim N(\mu(\bm{x}),\sigma^2(\bm{x}))$, the expected improvement (EI) can be expressed as below~\cite{Lyu2017An}
\begin{equation}
    \label{eq:EI}
    \begin{split}
	& EI(\bm{x}) = \sigma(\bm{x}) (\lambda * \text{CDF}(\lambda) + \text{PDF}(\lambda)),
    \end{split}
\end{equation}
where
$$
\lambda = \frac{\tau - \mu(\bm{x})}{\sigma(\bm{x})}
$$
In (\ref{eq:EI}), $\text{CDF}(.)$ and $\text{PDF}(.)$ are the CDF and PDF of the standard normal distribution respectively. $\mu(\bm{x})$ is the predictive mean at $\bm{x}$, and $\sigma(\bm{x})$ is the uncertainty estimation at input $\bm{x}$. The expected improvement function encourages both exploration and exploitation because it is large when the prediction is low (exploitation) or uncertainty is high (exploration)~\cite{Lyu2017An}.

%% wEI
The weighted expected improvement (wEI) criterion~\cite{gelbart2014bayesian} can work to encourage a minimized objective function value and legitimate constraints output. The formulation can be described as below
\begin{equation}
    \label{eq:wEI}
    wEI(\bm{x}) = EI(\bm{x}) \prod_{i=1}^{N_c} PF_{c}(\bm{x}),
\end{equation}
where $N_c$ denotes the number of constraints, and $PF_{i}(\bm{x})$ represents how promising the $i$-th constraint will be satisfied. The equation (\ref{eq:wEI}) works as a whole to favor the regions with high probability to be valid, and to reduce the acquisition function in regions where the constraints are likely to be violated. Thus, wEI can be used to propose new candidate recipe such that the objective function is minimized and the constraints are satisfied.

%%%%%%%%%%%%%%%%% log EI and log PI %%%%%%%%%%%%%%%%%%%%%%%%%%%%%%%%%%%%%%%%%%%%%%%%%%%%%%%%%%%%%%%%%%
%It is worth notice that when the number of constraints are large, the wEI will naturally be small even if every $PI_{c}(x)$ is around 0.99, which may be a great hinder in optimizing the input $x$. In order to overcome this numerical problem, we use $log(wEI(x))$ in this paper instead, where $log(.)$ denotes the logarithm function. As a results, the acquisition function is as follows:
%\begin{equation}
%    log(wEI(x)) = log(EI(x)) + \sum_{c}^{C} log(PI_{c}(x))
%\end{equation}
%The problem is that when $EI(x)$ or $PI_{c}(x)$ is too small, there is no way to calculate the logarithm function.
%\begin{equation}
%    log(EI(x)) = log(\sigma(x)) + log(\phi(\lambda)) + log(1+\frac{\lambda \Phi(\lambda)}{\phi(\lambda)})
%\end{equation}
%%%%%%%%%%%%%%%%%%%%%%%%%%%%%%%%%%%%%%%%%%%%%%%%%%%%%%%%%%%%%%%%%%%%%%%%%%%%%%%%%%%%%%%%%%%%%%%%%%%%%%

\section{Bayesian Optimization Using Neural Network}

Bayesian optimization has been demonstrated to be effective for analog circuit synthesis\cite{Lyu2017An}, however, the Gaussian process with pre-defined stationary kernel may not be able to properly characterize the data, in this work, we combine recent advancements of Gaussian process from machine learning community\cite{snoek2015scalable, huang2015scalable}, and propose to use neural-network-based Bayesian optimization for efficient analog circuit synthesis. On one hand, the kernel function of Gaussian process can be viewed as inner product of nonlinear feature maps, on the other hand, deep neural networks are famous for providing high-quality feature representations, with the above two properties, we can use deep neural network to \emph{automatically learn} a kernel function of Gaussian process.

\subsection{Gaussian Process Regression Model with Neural Network}

A Gaussian process model can also be derived from a weight space view. Denote $\bm{\phi}(\bm{x}): R^D \rightarrow R^M$ a feature map from $D$-dimension input space $\bm{x}$ to the $M$-dimension feature space. Assume that the function $f(\bm{x})$ can be expressed as a linear combination of nonlinear features. For the observed value $y$, we assume it differs from the function $f(\bm{x})$ by the additive noise. We assume the noise term follows the distribution of $N(0,\sigma^{2}_{n})$. $f(\bm{x})$ and $y$ can be expressed as
\begin{equation}
    \begin{cases}
        f(\bm{x}) = \bm{w}^{T} \bm{\phi}(\bm{x}) \\
        y-f(\bm{x}) \sim N(0,\sigma^{2}_{n}),
    \end{cases}
\end{equation}
where $\bm{w}\in R^M$ is the weight vector to be determined. If we impose a Gaussian prior with zero mean and covariance matrix $\Sigma_{p} = \frac{\sigma_p^2}{M}I \in R^{M \times M}$ on the weights $\bm{w}$, i.e., $\bm{w} \sim N(0, \Sigma_{p})$, it can be proved that $f$ follows a Gaussian process $f \sim \mathcal{GP}(0, k)$~\cite{rasmussen2004gaussian}, where the kernel function $k$ can be defined as
\begin{equation}
    \label{eq:kernel}
    k(\bm{x}_1,\bm{x}_2) = \bm{\phi}(\bm{x}_1)^{T} \Sigma_{p} \bm{\phi}(\bm{x}_2).
\end{equation}

For a new point $\bm{x}$, the prediction of $y$ given the training set $X, \bm{y}$ can be expressed as $P(y|\bm{x}, X, \bm{y}) \sim N(\mu(\bm{x}), \sigma^2(\bm{x})$, and
\begin{equation}
    \label{eq:nn_mu_and_sigma}
    \left\{
    \begin{array}{lll}
        \mu(\bm{x})        &=& \bm{\phi}(\bm{x})^{T} A^{-1} \Phi \bm{y} \\
        \sigma^{2}(\bm{x}) &=& \sigma^{2}_{n} + \sigma^{2}_{n} \bm{\phi}(\bm{x})^{T} A^{-1} \bm{\phi}(\bm{x}) \\
        \Phi               &=& (\bm{\phi}(\bm{x}_1),\bm{\phi}(\bm{x}_{2}),~\dots~,\bm{\phi}(\bm{x}_{N})) \\
        A                  &=& \Phi \Phi^{T} + \frac{M \sigma^{2}_{n}}{\sigma^{2}_{p}}I. \\
    \end{array}
    \right.
\end{equation}

\begin{figure*}
    \centering
    \includegraphics[width=0.7\textwidth]{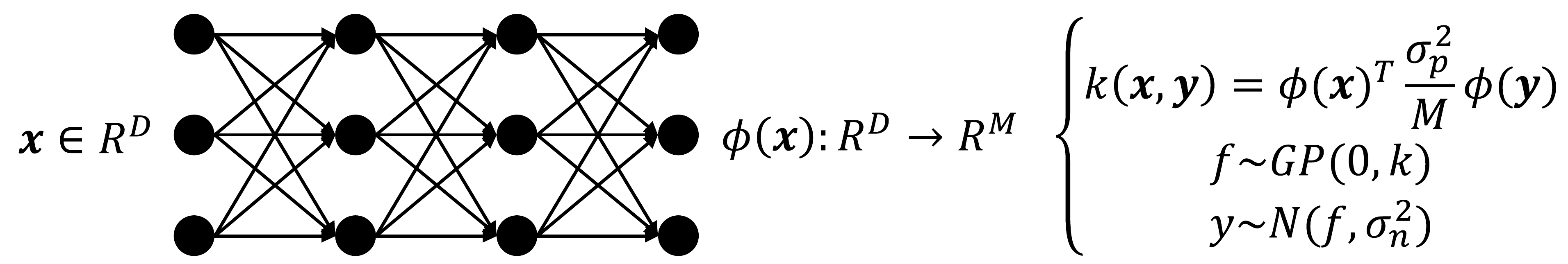}
    \caption{The Overall Architecture of the GP Model with Kernel Characterized by Neural Network}
    \label{fig:overall_architecture}
\end{figure*}

The logarithmic likelihood of the training data defined in equation (\ref{eq:MLE}) can also reformulated as follows\cite{huang2015scalable}
\begin{equation}
    \label{eq:nn_MLE}
    \begin{split}
        \log p(\bm{y}|X,\theta) = -\frac{1}{2 \sigma^{2}_{n}}(\bm{y}^{T}\bm{y} - \bm{y}^{T} \Phi^{T} A^{-1} \Phi \bm{y}) \\
         - \frac{1}{2} \log \lvert A \rvert + \frac{M}{2} \log \frac{M \sigma^{2}_{n}}{\sigma^{2}_{p}} - \frac{N}{2} \log(2 \pi \sigma^{2}_{n}),
    \end{split}
\end{equation}
where $\bm{\theta}$ is the vector that contains the hyper parameters $\sigma^{2}_{n}$, $\sigma^{2}_{p}$ and the hyper parameters of the feature map $\phi(.)$.

We introduce a neural network to implement the feature map. The architecture of the neural network is shown in Fig. \ref{fig:overall_architecture}. The neural network consists of 4 fully-connected layers including a input layer, 2 hidden layers and a output layer. ReLU function is taken as activation function in our neural network. The weights of the neural network can be viewed as the hyper parameters of the feature map, and the weights can be learned through a back propagation process as described in the next subsection. Neural-network-based GPR model provides more flexibility to characterize the similarities of the data points, and are thus possible to provide more accurate predictions and thus accelerate the follow-up optimization procedure.

%Obviously, the GP model defined from finite feature map reduces the computation time from cubic to linear in the number of training samples. The reduced-rank Gaussian process model not only significantly reduces the time to build a GP model, but also helps to lower the predictive operations, which in return decreases the cost of acquisition function evaluation. This allows us to directly balance between evaluation time and model capacity without the loss of flexibility and well-calibrated uncertainty.

\subsection{Neural Network Training}
%Before fully explain the training process of the GP model with neural network, the standard back-propagation algorithm will be presented\cite{bishop1995neural}. Let $f_{\theta}(x)$ denote the surrogate model, where $\theta$ represents the parameter vector and $x$ indicates the input vector. Considering the squared-error loss function, the corresponding gradient descent algorithm is as follows:
%\begin{equation}
%    \begin{cases}
%        y = f_{\theta}(x), \\
%        loss = \frac{1}{2} (t - y)^{T}(t - y),\\
%        \frac{\partial loss}{\partial \theta}  = (y - t) \frac{\partial f_{theta}(x)}{\partial \theta},\\
%        \theta_{new} = \theta - \alpha \frac{\partial loss}{\partial \theta},
%    \end{cases}
%\end{equation}
%where $t$ denotes the real value provided by the unknown objective function, and $\alpha$ denotes the learning rate. And $\theta_{new}$ represents the updated parameter vector, which will involved in the next iteration.

Denote $\bm{\eta}$ the weights of the neural network. The hyper parameters of the GPR model using neural network can be expressed as
$$
\bm{\theta} = [\sigma_{n}, \sigma_{p}, \bm{\eta}^{T}].
$$

We minimize the negative logarithmic likelihood in (\ref{eq:nn_MLE}) to obtain the hyper parameters $\bm{\theta}$. The minimization can be achieved through a gradient decent procedure. Denote $loss(\bm{\theta}) = - \log p(\bm{y}|X,\bm{\theta})$. The gradients of $loss(\bm{\theta})$ with respect to $\sigma_n$ and $\sigma_{p}$ can be obtained directly. We consider the gradient of $loss(\bm{\theta})$ with respect to the weights of the neural network $\bm{\eta}$. It can be obtained through the back propagation
\begin{equation} \label{chain}
\frac{\partial loss}{\partial \bm{\eta}} = \frac{\partial loss}{\partial \phi} \cdot \frac{\partial \phi}{\partial \bm{\eta}}.
\end{equation}
$\frac{\partial \phi}{\partial \bm{\eta}}$ is the gradient of the output of neural network with respect to the weights $\bm{\eta}$. From (\ref{chain}), we can find that the training of the neural network is actually embedded in the optimization procedure of maximizing the logarithmic likelihood (\ref{eq:nn_MLE}).

\subsection{Model Ensemble}
It has long been observed that randomly initializing a set of models and training each model with the same dataset can help to average out the random fluctuations and generate a more powerful model~\cite{Dietterich2000Ensemble}. The ensemble members work as a whole to improve the stability and accuracy of the prediction.

First, we construct $K$ independent probabilistic models with random initializations of the hyper parameters. Each model would give prediction distribution $p(y|\bm{x},\theta_{k}) = N(\mu_{k}(\bm{x}),\sigma^{2}_{k}(\bm{x}))$, where $\theta_{k}$ is the parameter vector of $k$-th model, and $\mu_{k}(\bm{x})$ and $\sigma^{2}_{k}(\bm{x})$ is the corresponding mean and variance. Then, the predictive distribution of combined model can expressed as $p(y|\bm{x})=N(\mu(\bm{x}),\sigma^{2}(\bm{x}))$, where:
\begin{equation}
    \label{eq:ensembles}
    \begin{cases}
	\mu(\bm{x}) = \frac{1}{K} \sum_{k}^{K} \mu_{k}(\bm{x})\\
	\sigma^{2}(\bm{x}) = \frac{1}{K} \sum_{k}^{K} (\mu^{2}_{k}(\bm{x}) + \sigma^{2}_{k}(\bm{x}))-\mu^{2}(\bm{x}).
    \end{cases}
\end{equation}

Recent research demonstrated that ensemble can greatly improve the quality of predicted uncertainty, and the performance will be enhanced especially for the data point which is far from the training set~\cite{lakshminarayanan2017simple}. The number of the ensemble members in this paper is empirically set to be 5. Note that the set of 5 probabilistic surrogate models can be constructed in parallel.

\subsection{Computational Complexity Analysis}
In (\ref{eq:mu_and_sigma}), the complexity of calculating $\mu(\bm{x})$ and $\sigma^{2}(\bm{x})$ is $O(N)$ and $O(N^2)$, respectively, as long as $(K+\sigma^{2}_{n}I)^{-1}$ and $(K+\sigma^{2}_{n}I)^{-1}(\bm{y}-\bm{m})$ are pre-calculated. In the GPR model with neural network, the complexity of calculating $\mu(\bm{x})$ and $\sigma^{2}(\bm{x})$ is $O(M)$ and $O(M^2)$, respectively, as long as $A^{-1}$ and $A^{-1} \Phi \bm{y}$ are pre-calculated. Note that matrix $A$ is a $M\times M$ matrix.

For the training of the model, the complexity of evaluating (\ref{eq:MLE}) is limited by the inverse of the $K_{\theta}$, and the complexity of the inverse of $K_{\theta}$ is $O(N^3)$. For GPR model with neural network, the complexity of evaluating the the negative logarithmic likelihood in (\ref{eq:nn_MLE}) is limited by the inverse of $A$ and $(\Phi \bm{y})^{T}A^{-1}(\Phi \bm{y})$. The complexity of inverse $A$ is $O(M^3)$, and the complexity of calculating $(\Phi \bm{y})^{T}A^{-1}(\Phi \bm{y})$ is $O(M^3 +NM^2)$. Thus, the complexity of evaluating (\ref{eq:nn_MLE}) is $O(M^3 +NM^2)$.

Note that $M$ is fixed while the $N$ is the number of training points and $N$ will be increased gradually during the optimization procedure. From the complexity analysis, we can see that the traditional GPR model scales cubically with the number of observed data points $N$, which can be a great challenge for objective functions that require a large number of iterations. On the other hand, our proposed GPR model with neural network scales linearly with the number of observed data points $N$. It is more suitable for the applications that requires a large number of iterations.

\subsection{Summary}
\begin{figure}[tbp]
    \centering
    \includegraphics[width=0.38\textwidth]{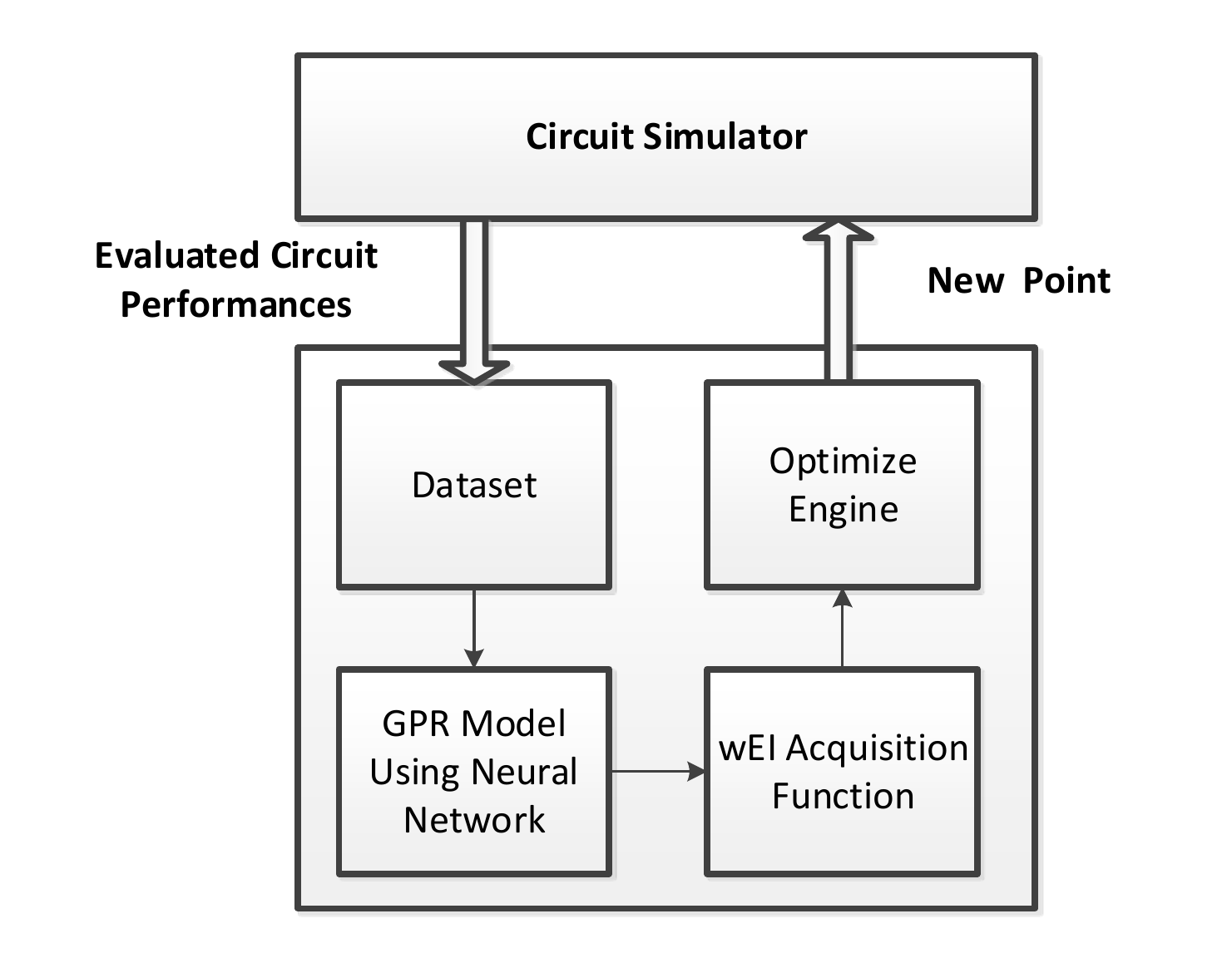}
    \caption{The Proposed Bayesian Optimization Framework}
    \label{fig:summary}
\end{figure}

The proposed GP model-based Bayesian optimization approach is summarized in Algorithm \ref{algo:Bayesian_optimization}. The framework of the algorithm is also shown in Fig. \ref{fig:summary}. Firstly, we randomly generate a set of training data points. Based on the initial training set, the GPR model using neural network is built. Model ensemble is used to improve the quality of predicted uncertainty. Then, we maximize the acquisition function to find the new design point $\bm{x}_t$. The performances at $\bm{x}_t$ is simulated and added to the training set. A new model is built and the optimization continues until convergence or the simulation budget is exhausted.

\begin{algorithm}
    \caption{Constrained single-objective Bayesian optimization algorithm}
    \label{algo:Bayesian_optimization}
    \begin{algorithmic}
        \STATE Initialize a set of training data $D = \{X,\bm{p}\}$, where $\bm{p}$ represents a set of performances
        \FOR{t = 1 to N}
        \FOR{s = 1 to 5}
        \STATE Randomly initializing the hyper parameters
        \STATE Training the GPR model using neural network by maximizing the logarithmic likelihood (\ref{eq:nn_MLE})
        \ENDFOR
        \STATE Model ensemble according to (\ref{eq:ensembles})
	\STATE Find $\bm{x}_t$ that maximize the acquisition function (\ref{eq:wEI}) using the predictions in (\ref{eq:ensembles})
	\STATE Simulate the design at $\bm{x}_t$ to get the performances $\bm{p}_t$
	\STATE Add the point $(\bm{x}_t, \bm{p}_t)$ to the training set
        \ENDFOR
    \end{algorithmic}
\end{algorithm}

\section{Experimental Results}
In this section, we will demonstrate the effectiveness of the proposed method using two real-world analog circuits. We compared our proposed approach with WEIBO algorithm~\cite{Lyu2017An}, GASPAD algorithm~\cite{Liu2014GASPAD} and differential evolution (DE) algorithm~\cite{liu2009analog}. WEIBO is a GP-based Bayesian optimization approach with wEI works as the acquisition function. The GASPAD methodology is also a traditional GP-based approach with the evolutionary algorithm works as the optimization engine. DE is based on the evolutionary algorithm. Our experiments were conducted on a Linux workstation with two Intel Xeon CPUs and 128G memory. HSPICE is used for circuit simulation.

\subsection{A Two-Stage Operational Amplifier}

\begin{figure}
    \centering
    \includegraphics[width=0.40\textwidth]{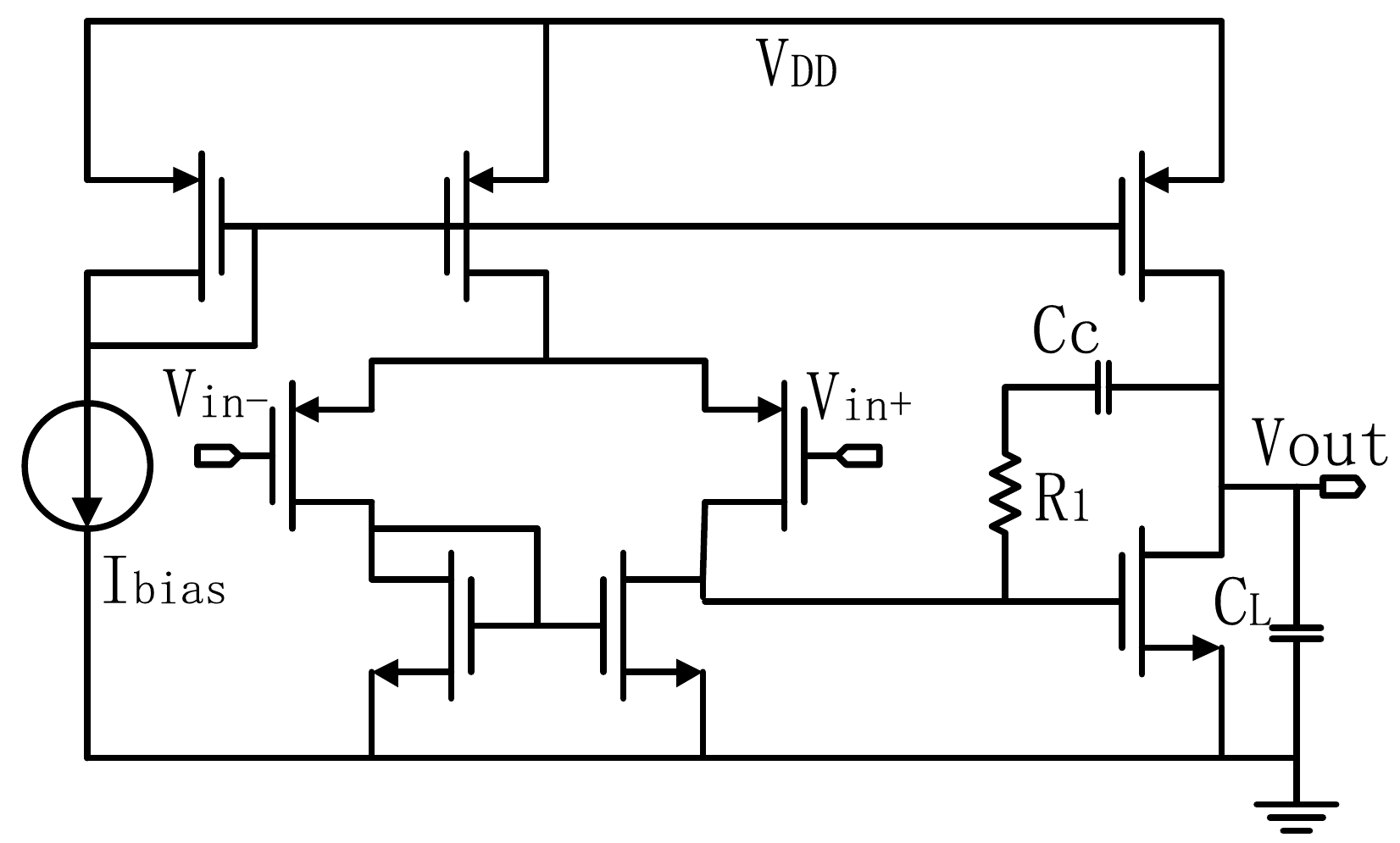}
    \caption{An two-stage operational amplifier~\cite{Yang2018Smart}.}
    \label{fig:two_stage_opamp}
\end{figure}

The first test circuit is a two-stage operational amplifier as shown in Fig. \ref{fig:two_stage_opamp}. This circuit is designed using a SMIC 180nm process. This circuit has 10 design variables, including the lengths and widths of the transistors. The corresponding design specifications are listed as follows.
\begin{equation}
    \label{eq:two_stage_opamp_spec}
    \begin{aligned}
        \text{maximize} \quad & GAIN & & \\
        \text{s.t.}     \quad & UGF  \quad & > \quad & 40MHz \\
                              & PM   \quad & > \quad & 60^{o},
    \end{aligned}
\end{equation}
where GAIN represents the open-loop gain, UGF means the unity gain frequency, and PM denotes the phase margin~\cite{Yang2018Smart}. We ran the optimization algorithms 10 times to average out the random fluctuations. The number of initial samples is set to be 30. The maximum number of simulations is limited to 100 for both the proposed algorithm and WEIBO. For GASPAD and DE, the corresponding maximum number of simulations is set to be 200 and 1100. The experimental results are shown in Table \ref{table:two_stage_opamp}. Compared with WEIBO, our proposed algorithm reduced the simulation time on average by 7\% while finding better design. Although both GASPAD and DE can find the feasible point, the design they find are much worse while consuming more number of simulations.

\begin{table}
    \centering
    \caption{The optimization results of the two-stage opamp.}
    \label{table:two_stage_opamp}
    \begin{tabular}{ccccc}
        \hline
        \textbf{Alg} & \textbf{Ours} & \textbf{WEIBO} & \textbf{GASPAD} & \textbf{DE}\\
        \hline
        UGF & 40.08 & 42.67 & 40.70 & 40.25 \\
        PM & 61.33 & 61.03 & 62.79 & 61.34 \\
        \hline
        mean & \textbf{88.17} & 87.95 & 85.90 & 87.57 \\
        median & \textbf{88.29} & 87.77 & 86.34 & 87.59 \\
        best & \textbf{89.95} & 89.90 & 87.27 & 88.53 \\
        worst & \textbf{86.00} & 85.39 & 83.52 & 86.34\\
        \hline
        Avg. \# Sim & 86 & 92 & 122 & 999 \\
        \# Success & 10/10 & 10/10 & 10/10 & 10/10 \\
        \hline
    \end{tabular}
\end{table}

\subsection{A Charge Pump}

\begin{figure*}
    \centering
    \includegraphics[width=0.65\textwidth]{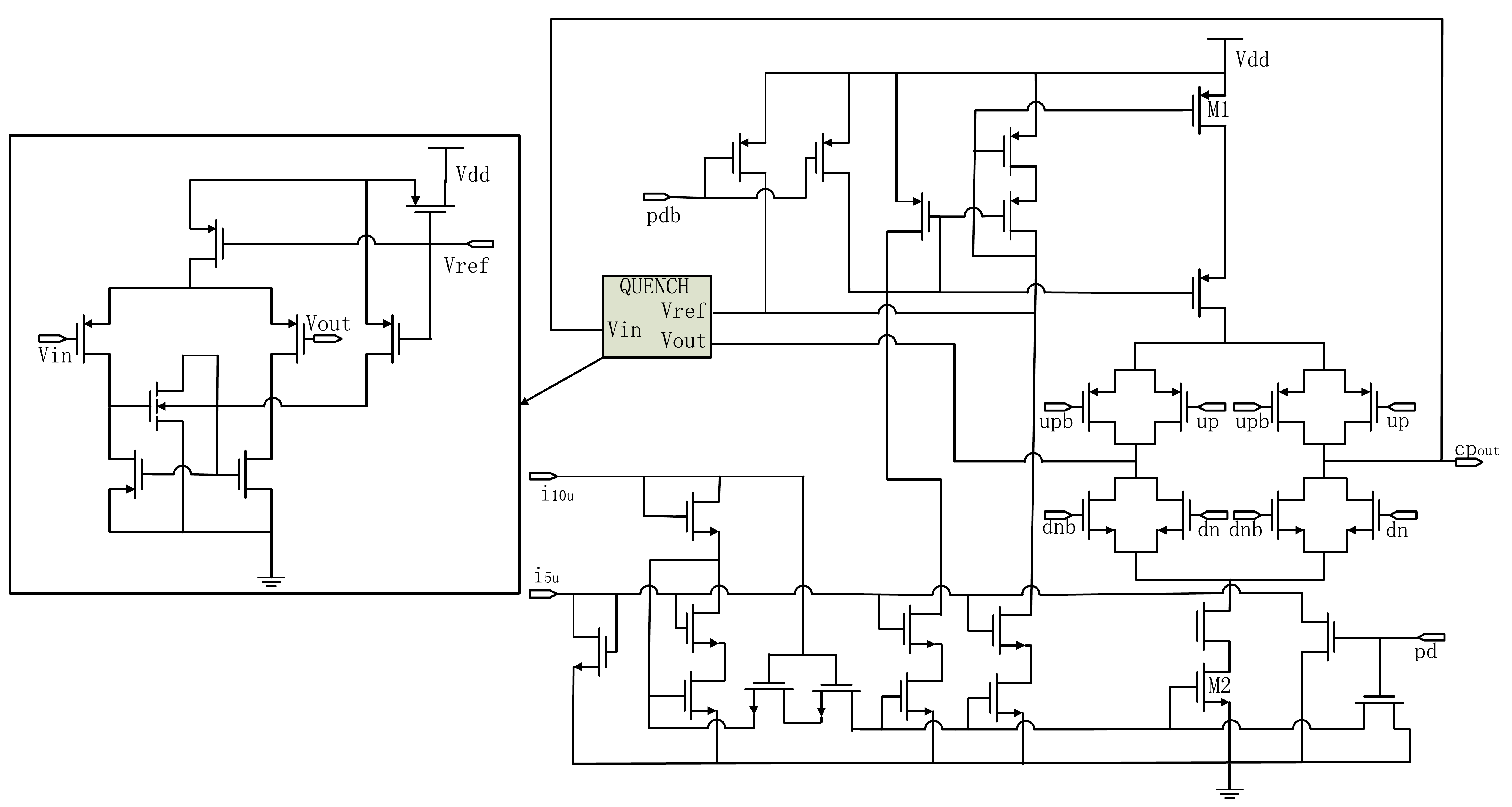}
    \caption{Schematic of the Charge Pump Circuit, Schematic Reproduced with Permission from~\cite{Lyu2017An}}
    \label{fig:charge_pump}
\end{figure*}

The second test circuit is a charge pump circuit implemented in a SMIC 40nm process. This circuit is provided by the authors of~\cite{Lyu2017An}, which is the same as the charge pump test case used in~\cite{Lyu2017An}. The schematic of the circuit is shown in Fig. \ref{fig:charge_pump}. There are 36 design variables in this test case. In this test case, we take a total of 18 PVT corners into consideration. The corresponding design specifications are listed as follows.
\begin{equation}
    \begin{aligned}
        \text{minimize} \quad & FOM & & \\
        \text{s.t.} \quad & \mathit{diff_{1}} \quad & < \quad & 20\mu A, \\
             & \mathit{diff_{2}} \quad & < \quad & 20 \mu A, \\
             & \mathit{diff_{3}} \quad & < \quad & 5 \mu A, \\
             & \mathit{diff_{4}} \quad & < \quad & 5 \mu A, \\
             & deviation \quad & < \quad & 5 \mu A, \\
    \end{aligned}
\end{equation}
where
\begin{equation}
    \begin{cases}
        \mathit{diff_{1}} = max_{\forall PVT}(I_{M1,max} - I_{M1,avg}), \\
        \mathit{diff_{2}} = max_{\forall PVT}(I_{M1,avg} - I_{M1,min}), \\
        \mathit{diff_{3}} = max_{\forall PVT}(I_{M2,max} - I_{M2,avg}), \\
        \mathit{diff_{4}} = max_{\forall PVT}(I_{M2,avg} - I_{M2,min}), \\
	\mathit{diff} = \sum_{i=1}^{4} \mathit{diff_{i}}, \\
        deviation = max_{\forall PVT}(\lvert I_{M1,avg} - 40\mu A \rvert) \\
        + max_{\forall PVT}(\lvert I_{M2,avg} - 40 \mu A\rvert),\\
	FOM = 0.3 \times \mathit{diff} + 0.5 \times deviation.
    \end{cases}
\end{equation}

In this experiment, the number of initial samples is 100, and the maximum number of simulations is 790. The number of ensemble members is empirically set to be 5. We ran the optimization algorithms for 12 times to average out the random fluctuations. The results of the optimization are shown in Table \ref{table:charge_pump}. Compared with WEIBO, we reduced approximately 28.9\% number of the simulations while the objective function exceed WEIBO by 0.68 even for the worst case. Both GASPAD and DE algorithm find feasible solution in the experiment. However, the designs they find after 2000 simulations are much worse than our algorithm. We can see from Table \ref{table:charge_pump} that GP model with neural network can help to achieve much better results with less number of simulations.

\begin{table}
    \centering
    \caption{The Optimization results of the charge pump, the results of WEIBO, GASPAD and DE come from \cite{Lyu2017An}}
    \label{table:charge_pump}
    \begin{tabular}{ccccc}
        \hline
        \textbf{Algo} & \textbf{Ours} & \textbf{WEIBO} & \textbf{GASPAD} & \textbf{DE} \\
        \hline
        $\rm diff_{1}$ & 5.69 & 6.58 & 6.83 & 17.97 \\
        $\rm diff_{2}$ & 4.30 & 5.30 & 5.28 & 15.49 \\
        $\rm diff_{3}$ & 0.13 & 0.24 & 0.29 & 1.84 \\
        $\rm diff_{4}$ & 0.18 & 0.37 & 0.40 & 3.56 \\
        deviation & 0.15 & 0.41 & 0.33 & 0.39 \\
        \hline
        mean & \textbf{3.48} & 3.95 & 4.00 & 11.85 \\
        median & \textbf{3.36} & 3.97 & 4.99 & 12.31 \\
        best & \textbf{3.17} & 3.48 & 3.74 & 9.29 \\
        worst & \textbf{3.80} & 4.48 & 4.43 & 13.40 \\
        \hline
        Avg. \# Sim & \textbf{562} & 790 & 2328 & 1538 \\
        \# Success & \textbf{12/12} & 12/12 & 12/12 & 12/12 \\
        \hline
    \end{tabular}
\end{table}

%%%%%%%%%%%%%%%%%%%%%%%%%%%%%%%%%%%%%%%%%%%%
%\begin{table}
%    \centering
%    \caption{structure of the neural network for two-stage opamp}
%    \label{table:nn_two_stage_opamp}
%    \begin{tabular}{cccc}
%        \hline
%        \textbf{output index} & \textbf{0} & \textbf{1} & \textbf{2} \\
%        \hline
%        \textbf{number of layers} & 1 & 1 & 1 \\
%        \hline
%        \textbf{size of each layer} & 10 & 10 & 10 \\
%        \hline
%        \textbf{activation function} & relu & relu & relu \\
%        \hline
%    \end{tabular}
%\end{table}
%%%%%%%%%%%%%%%%%%%%%%%%%%%%%%%%%%%%%%%%%%%%%%

%%%%%%%%%%%%%%%%%%%%%%%%%%%%%%%%%%%%%%%%%%%%%%%%%%%
%\begin{table}
%    \centering
%    \caption{structure of the neural network for charge pump}
%    \label{table:charge_pump_nn_structure}
%    \begin{tabular}{ccccccc}
%        \hline
%        \textbf{output index} & \textbf{0} & \textbf{1} & \textbf{2} & \textbf{3} & \textbf{4} & \textbf{5} \\
%        \hline
%        \textbf{number of layers} & 2 & 2 & 2 & 2 & 2 & 2 \\
%        \hline
%        \textbf{size of each layer} & 40 & 20 & 10 & 20 & 25 & 10 \\
%        \hline
%        \textbf{activation function} & relu & relu & relu & relu & relu & relu \\
%        \hline
%    \end{tabular}
%\end{table}
%%%%%%%%%%%%%%%%%%%%%%%%%%%%%%%%%%%%%%%%%%%%%%%%%%%

\section{Conclusion}
In this paper, we proposed a Bayesian optimization approach for analog circuit synthesis using neural network. A well-designed neural network is proposed to provide efficient feature mapping. Model ensemble is proposed to improve the quality of uncertainty prediction. Compared to Gaussian process regression model with explicitly defined kernel functions, neural-network-based models are possible to provide more accurate predictions and thus accelerate the follow-up optimization procedure. Experimental results of the real-world analog circuit synthesis problems show that our proposed approach achieves better optimization results with less number of simulations than the state-of-the-art algorithm. Moreover, our proposed GPR model with neural network scales linearly with the number of observed data points $N$. It is more suitable for the applications that requires a large number of iterations. 

\section{Acknowledgement}
This research is supported partly by the National Major Science
and Technology Special Project of China (2017ZX01028101-003), partly by National Natural Science Foundation
of China (NSFC) research projects 61822402, 61574046, 61774045, 61674042
 61574044, 61628402, and 61474026.

%\section*{Acknowledgment}

%\section*{References}

\bibliographystyle{unsrt}
\bibliography{ref}
%\printbibliography

\end{document}